\documentclass[conference]{IEEEtran}
\usepackage{cite}
\usepackage{amsmath,amssymb,amsfonts}
\usepackage{algorithmic}
\usepackage{graphicx}
\usepackage{textcomp}
\usepackage{xcolor}
\usepackage{hyperref}
\usepackage{makecell}
\usepackage{placeins}
\usepackage{subcaption}

\begin{document}

\title{Beyond CLIP: Knowledge-Enhanced Multimodal Transformers for Cross-Modal Alignment in Diabetic Retinopathy Diagnosis}

\author{\IEEEauthorblockN{Argha Kamal Samanta}
\IEEEauthorblockA{
\textit{Department of Electronics and}\\
\textit{Electrical Communication Engineering}\\
\textit{Indian Institute of Technology, Kharagpur}\\
Kharagpur, India\\
arghakamal25@gmail.com
}
\and
\IEEEauthorblockN{Harshika Goyal}
\IEEEauthorblockA{
\textit{Department of Electronics and}\\
\textit{Electrical Communication Engineering}\\
\textit{Indian Institute of Technology, Kharagpur}\\
Kharagpur, India\\
goyalharshika266@gmail.com
}
\and
\IEEEauthorblockN{Vasudha Joshi}
\IEEEauthorblockA{
\textit{Department of Computer}\\
\textit{Science and Engineering}\\
\textit{Indian Institute of Technology, Kharagpur}\\
Kharagpur, India\\
vasudhajoshi9@iitkgp.ac.in
}
\and
\IEEEauthorblockN{Tushar Mungle}
\IEEEauthorblockA{\textit{Department of Medicine} \\
\textit{Stanford University}\\
Stanford, USA \\
tushar.mungle@gmail.com}
\and
\IEEEauthorblockN{Pabitra Mitra}
\IEEEauthorblockA{
\textit{Department of Computer}\\
\textit{Science and Engineering}\\
\textit{Indian Institute of Technology, Kharagpur}\\
Kharagpur, India\\
pabitra@cse.iitkgp.ac.in
}
}

\maketitle

\begin{abstract}
Diabetic retinopathy (DR) is a leading cause of preventable blindness worldwide, necessitating accurate automated diagnostic systems. While general-domain vision-language models like Contrastive Language-Image Pre-Training (CLIP) have demonstrated impressive capabilities in natural image tasks, they exhibit severe limitations in medical domain applications, particularly in cross-modal retrieval for ophthalmological images. This paper presents a novel knowledge-enhanced joint embedding framework that integrates retinal fundus images, clinical text, and structured patient data through a multimodal transformer architecture to address the critical gap in medical image-text alignment.

Our approach employs separate encoders for each modality: a Vision Transformer (ViT-B/16) for retinal images, Bio-ClinicalBERT for clinical narratives, and a multilayer perceptron for structured demographic and clinical features. These modalities are fused through a joint transformer with modality-specific embeddings, trained using multiple objectives including contrastive losses between modality pairs, reconstruction losses for images and text, and classification losses for DR severity grading according to ICDR and SDRG schemes.

Experimental results on the Brazilian Multilabel Ophthalmological Dataset (BRSET) demonstrate significant improvements over baseline models. Our framework achieves near-perfect text-to-image retrieval performance with Recall@1 of 99.94\% compared to fine-tuned CLIP's 1.29\%, while maintaining state-of-the-art classification accuracy of 97.05\% for SDRG and 97.97\% for ICDR. Furthermore, zero-shot evaluation on the unseen DeepEyeNet dataset validates strong generalizability with 93.95\% Recall@1 versus 0.22\% for fine-tuned CLIP. These results demonstrate that our multimodal training approach effectively captures cross-modal relationships in the medical domain, establishing both superior retrieval capabilities and robust diagnostic performance.
\end{abstract}


\section{Introduction}
Diabetic retinopathy (DR) represents one of the most prevalent microvascular complications of diabetes mellitus and stands as a leading cause of preventable blindness in working-age adults globally. With the International Diabetes Federation projecting the number of people with diabetes to reach 700 million by 2045, the burden of DR-related vision impairment continues to escalate dramatically. Early detection and timely intervention are crucial for preventing irreversible vision loss, yet traditional screening methods face significant challenges in scalability, accessibility, and cost-effectiveness.

Manual examination of retinal fundus images by trained ophthalmologists remains the gold standard for DR diagnosis. However, this approach is time-consuming, expensive, and subject to inter-observer variability. More critically, the shortage of specialized ophthalmologists, particularly in low- and middle-income countries where diabetes prevalence is rising fastest, creates a substantial gap between screening demand and available resources. These limitations have motivated the development of automated screening systems powered by artificial intelligence.

Recent advances in deep learning and computer vision have demonstrated remarkable success in automated medical image analysis. Vision-language models such as CLIP (Contrastive Language-Image Pre-training) have revolutionized multi-modal learning in natural image domains, achieving impressive zero-shot and few-shot capabilities by learning joint representations of images and text \cite{radford2021learningtransferablevisualmodels}. However, despite their success in general domains, these models face critical limitations when applied to specialized medical contexts, particularly in ophthalmology.

Our preliminary investigations revealed a striking performance gap: while CLIP maintains high classification accuracy on retinal images, it catastrophically fails at cross-modal retrieval tasks that are essential for clinical decision support. Specifically, CLIP without fine-tuning achieves 0\% Recall@1 in text-to-image retrieval for retinal images, and even after domain-specific fine-tuning, performance only reaches 1.29\%. This severe misalignment between clinical descriptions and retinal images fundamentally limits the practical utility of such models in real-world clinical workflows where physicians need to retrieve relevant images based on clinical descriptions, generate reports from images, and integrate multi-modal patient data for comprehensive diagnosis.

The fundamental challenge stems from the fact that medical images contain highly specialized visual features like microaneurysms, exudates, hemorrhages, neovascularization, that require domain-specific knowledge to align with corresponding clinical terminology. General-purpose vision-language models trained on web-scraped image-text pairs lack this medical domain knowledge and fail to capture the fine-grained pathological features critical for clinical understanding. Moreover, most existing DR detection systems focus exclusively on image-based analysis, neglecting the wealth of complementary information available in electronic health records, including structured clinical variables such as age, diabetes duration, insulin use, and patient demographics.

In this work, we propose a comprehensive knowledge-enhanced joint embedding framework that addresses these limitations by integrating three distinct modalities: retinal fundus images, clinical text narratives, and structured patient data. Our approach employs modality-specific , i.e., a Vision Transformer for images, Bio-ClinicalBERT for clinical text, and a multilayer perceptron for structured features, combined with a novel joint transformer fusion mechanism that enables cross-modal attention and contextualization \cite{dosovitskiy2020image, alsentzer2019publiclyavailableclinicalbert}. Through multi-objective training incorporating contrastive alignment losses, modality reconstruction losses, and supervised classification losses with learnable dynamic weighting, our model learns robust joint representations that capture complementary diagnostic signals across modalities.

Evaluated on the Brazilian Multilabel Ophthalmological Dataset (BRSET) containing 16,266 fundus images from 8,524 patients \cite{10.1371/journal.pdig.0000454}, our framework demonstrates dramatic improvements in cross-modal retrieval while maintaining state-of-the-art classification performance. Specifically, we achieve 99.94\% Recall@1 in text-to-image retrieval compared to fine-tuned CLIP's 1.29\%, while maintaining classification accuracies of 97.05\% for SDRG and 97.97\% for ICDR grading schemes. Furthermore, zero-shot evaluation on the completely unseen DeepEyeNet dataset validates strong generalizability with 93.95\% Recall@1, establishing the robustness of our learned representations \cite{roy2025deepeyenetadaptivegeneticbayesian}.

The main contributions of this paper are: (1) a unified multimodal architecture that effectively integrates heterogeneous medical data types through modality-aware fusion; (2) empirical demonstration that general-domain vision-language models fail critically at medical cross-modal alignment tasks; (3) a multi-task learning strategy with dynamic loss weighting that jointly optimizes alignment, reconstruction, and classification objectives; (4) comprehensive evaluation demonstrating better retrieval performance without sacrificing diagnostic accuracy; and (5) strong cross-dataset generalization validating the potential for real-world clinical deployment.

The remainder of this paper is organized as follows: Section II reviews related work in deep learning for DR detection, vision-language models, and multi-modal learning in healthcare. Section III describes the methodology including data preprocessing and model architecture. Section IV presents experimental results on retrieval and classification tasks. Section V concludes with discussion of findings and future directions.

\section{Related Work}

\subsection{Deep Learning for Diabetic Retinopathy Detection}

The application of deep learning to automated DR screening has evolved significantly over the past decade. Early pioneering work demonstrated that deep learning algorithms could achieve sensitivity and specificity comparable to human experts in detecting DR from retinal fundus photographs \cite{10.1001/jama.2016.17216}. Recent systems like DeepDR have advanced the field by incorporating real-time image quality assessment, lesion detection, and multi-grade classification, achieving area under the curve values ranging from 0.901 to 0.972 across different DR severity levels \cite{Dai2021DeepDR}.

Convolutional neural networks, particularly architectures based on ResNet, DenseNet, Inception, and EfficientNet, have been extensively studied for DR classification tasks \cite{PRATT2016200}. Transfer learning from ImageNet has proven particularly effective for addressing limited medical imaging data, with models achieving classification accuracies exceeding 97\% on benchmark datasets. Advanced activation functions and data augmentation strategies have further improved performance, with recent approaches reporting accuracy as high as 99.41\% on specialized datasets \cite{healthcare11010097}. However, most existing approaches focus exclusively on image-based classification, neglecting complementary clinical information.

\subsection{Vision Transformers in Medical Imaging}

The remarkable performance of transformer architectures in natural language processing has triggered broad interest in computer vision, with transformers demonstrating superior capability in learning long-range dependencies and spatial correlations compared to traditional CNNs \cite{dosovitskiy2020image}. Vision transformers utilize self-attention mechanisms to capture intricate patterns crucial for medical image classification, with studies reporting substantial improvements over CNN-based approaches across diverse medical imaging modalities \cite{Takahashi2024}.

In medical imaging specifically, transformers offer appealing features including better scalability, enhanced robustness to image corruption, and weak inductive bias that enables superior performance when combined with large-scale datasets \cite{SHAMSHAD2023102802}. A systematic comparison of 36 studies indicates that transformer-based models, particularly vision transformers, exhibit significant potential in diverse medical imaging tasks, showcasing superior performance when contrasted with conventional CNN models \cite{article}. Hybrid architectures combining ViT with CNNs have emerged to mitigate the limitations of each architecture, offering comprehensive solutions that integrate global context understanding with precise local feature extraction \cite{djoumessi2025hybridfullyconvolutionalcnntransformer}.

For ophthalmological applications, vision transformers have demonstrated particular promise in capturing subtle pathological features distributed across retinal fundus images. Comprehensive reviews spanning multiple diseases including diabetic retinopathy confirm that ViT-based approaches often outperform traditional CNNs, particularly when leveraging transfer learning and domain adaptation techniques \cite{Aburass2025VisionTransformersMedImaging}.

\subsection{Vision-Language Models and Their Medical Domain Limitations}

CLIP introduced contrastive language-image pre-training as a paradigm for aligning image and text representations in a joint embedding space, demonstrating impressive zero-shot capabilities across various natural image domains \cite{radford2021learningtransferablevisualmodels}. The model jointly optimizes vision and text encoders using a contrastive objective that brings matching image-text pairs closer while pushing apart non-matching pairs in the embedding space.

However, applying CLIP to medical imaging presents unique challenges including data scarcity, complex cross-modal alignment requirements, and the need for domain-specific knowledge not captured in general web-scraped training data \cite{Zhao_2025}. Medical images contain highly specialized visual features that require domain-specific knowledge to align with corresponding clinical terminology, and CLIP's text encoder limitation of 77 tokens is insufficient for information-rich radiology reports \cite{Hartsock2024VLMMedical}. Recent work has shown that clinical text often exceeds several hundred tokens with detailed descriptions of anatomical structures and diagnostic impressions, which CLIP must truncate, resulting in inevitable information loss and degraded alignment \cite{lin2025tamingvisionlanguagemodelsmedical}.

Domain-specific adaptations have emerged, with models like Mammo-CLIP demonstrating that capitalizing on alignment between visual and textual data in specialized medical contexts can significantly improve performance in tasks such as abnormality detection \cite{chen2024mammoclipleveragingcontrastivelanguageimage}. Recent studies emphasize task-driven prompt optimization and text-parameterized convolution operations to enhance medical VLM adaptability, though challenges remain in achieving robust cross-modal alignment without massive domain-specific datasets \cite{xiao2025promptbasedadaptationlargescalevision}. Furthermore, vision-language foundation models have been shown to exhibit demographic biases in medical imaging, consistently underdiagnosing marginalized groups, raising important ethical considerations for clinical deployment \cite{Yang2025DemographicBiasVLM}.

\subsection{Multimodal Learning in Healthcare}

Modern clinical decision-making typically relies on integrating multiple data sources including demographic information, laboratory results, vital signs, and imaging data \cite{Patil2025MDSS}. Recent systematic reviews demonstrate that integrating structured and unstructured medical data significantly improves performance in diagnostic tasks, with multimodal approaches outperforming unimodal baselines in diagnosis, prognosis prediction, and personalized treatment planning \cite{article2}.

Multimodal fusion strategies can be categorized into early, late, and intermediate approaches. Early fusion combines raw features before processing, while late fusion processes each modality independently before combining predictions. Intermediate fusion through transformers has emerged as particularly effective, offering modality-specific feature extraction followed by learned cross-modal interactions through attention mechanisms \cite{SAGHEER20254259}. Graph neural networks have also gained interest for modeling non-Euclidean structures in multimodal healthcare data, explicitly capturing complex relationships between imaging features and clinical parameters \cite{Waqas2024MultimodalOncology}.

Multimodal large language models have shown promise in healthcare by processing diverse data types including medical images, time-series data, audio, text, and omics data \cite{info:doi/10.2196/59505}. The integration of heterogeneous data modalities through natural language prompting strategies has proven particularly effective for fusing diverse EHR data types without requiring conversion to standardized formats \cite{ren2025comprehensivesurveyelectronichealth}. Cross-attention transformer modules for merging features from heterogeneous modalities have demonstrated superior performance compared to traditional fusion approaches in multimodal medical diagnosis tasks \cite{10056308}.

\subsection{Contrastive Learning for Medical Imaging}

Contrastive learning has emerged as a powerful self-supervised paradigm that learns representations by bringing similar samples closer and pushing dissimilar samples apart in the embedding space. A systematic review of self-supervised learning in medical imaging from 2012 to 2022 revealed that contrastive learning strategies, particularly SimCLR, MoCo, and BYOL frameworks, were the most commonly adopted approaches, appearing in 44 of 79 reviewed studies \cite{Shurrab2022SelfSupervisedMedicalImaging}.

Local region contrastive learning approaches have shown particular promise in medical imaging by focusing on significant image regions and cross-modality interactions, with attention-weighted representations enabling context-aware feature learning \cite{NEURIPS2020_949686ec}. Federated contrastive learning enables collaborative model training across decentralized medical datasets while preserving privacy, with feature exchange during training providing diverse contrastive data for effective learning \cite{11040479}. Recent work emphasizes the importance of expert annotations and challenging negative mining in medical contrastive learning, with enhanced alignment and uniformity leading to improved zero-shot inference and cross-modal retrieval capabilities \cite{kumar2024improvingmedicalmultimodalcontrastive}.

Modern training strategies for multimodal tasks emphasize alignment and integration of features across different modalities, with contrastive learning on large-scale image-text pairs maximizing similarity of correct pairs while minimizing incorrect ones \cite{pmlr-v182-zhang22a}. Region-aware multimodal contrastive learning frameworks that incorporate both global and region-specific features have demonstrated enhanced accuracy and granularity in medical image-text alignment \cite{NEURIPS2020_949686ec}.

\subsection{Research Gap and Motivation}

Despite significant progress in both medical image analysis and vision-language modeling, critical gaps remain that limit the clinical applicability of existing approaches:

\textbf{Domain Mismatch:} General-domain vision-language models trained on web-scraped data fail catastrophically when applied to specialized medical contexts. Our preliminary experiments revealed that CLIP achieves 0\% Recall@1 on retinal image retrieval tasks without fine-tuning, and only 1.29\% even after domain-specific fine-tuning, demonstrating a fundamental inability to establish meaningful cross-modal alignments in ophthalmology.

\textbf{Unimodal Focus:} Most DR detection systems rely exclusively on fundus images, ignoring valuable structured clinical data including patient demographics, diabetes duration, insulin use, and comorbidities that clinicians routinely consider in diagnostic workflows.

\textbf{Retrieval Task Neglect:} While classification accuracy has improved substantially, cross-modal retrieval capabilities which are essential for clinical decision support systems that need to match images with descriptions or generate reports from images, remain largely unexplored and poorly performing.

\textbf{Limited Generalization:} Many models are developed and evaluated on single, homogeneous datasets, raising concerns about real-world applicability when deployed across different populations and clinical settings.

This work addresses these gaps by proposing a knowledge-enhanced multimodal transformer framework specifically designed for medical domain applications, integrating images, clinical text, and structured patient data through joint embedding learning with multi-objective training that simultaneously optimizes cross-modal alignment, reconstruction quality, and diagnostic classification.

\section{Methodology}

\subsection{Data Preprocessing}
The experiments were conducted using the Brazilian Multilabel Ophthalmological Dataset (BRSET), which contains 16,266 macula-centred colour fundus photographs from 8,524 patients collected across three ophthalmological centres in São Paulo, Brazil, between 2010 and 2020. Each image is accompanied by demographic, anatomical, and pathological annotations, including diabetic retinopathy severity graded according to both the International Clinical Diabetic Retinopathy (ICDR) and Scottish Diabetic Retinopathy Grading (SDRG) schemes.

\begin{figure}[htbp]
\centerline{\includegraphics[width=\columnwidth]{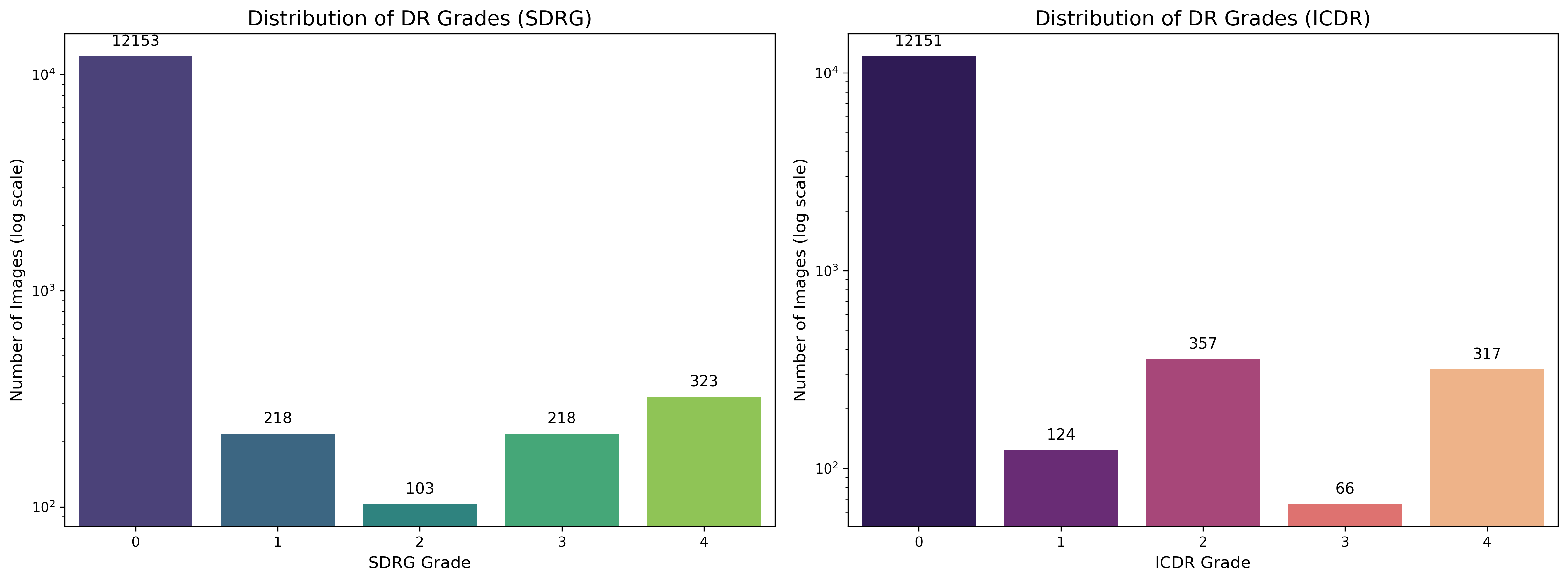}}
\caption{Distribution of DR grades in the BRSET training dataset.}
\label{fig:dr_grades}
\end{figure}

Fundus images were captured using Nikon NF505 and Canon CR-2 cameras, resized to $224 \times 224$ px, normalised to the $[0,1]$ range, and standardised with ImageNet statistics. To improve generalisation, random horizontal/vertical flips, $\pm15^\circ$ rotations, and mild brightness and contrast jitter were applied during training, while validation and test sets were left unaltered.

Structured clinical features were extracted, including age, sex, exam eye, duration of diabetes, insulin use, and diabetes diagnosis. Continuous variables were standardised to zero mean and unit variance, categorical attributes were numerically encoded, and missing values were imputed using the median or mode as appropriate.

\begin{figure}[htbp]
\centering

\begin{subfigure}{\columnwidth}
    \centering
    \includegraphics[width=\linewidth]{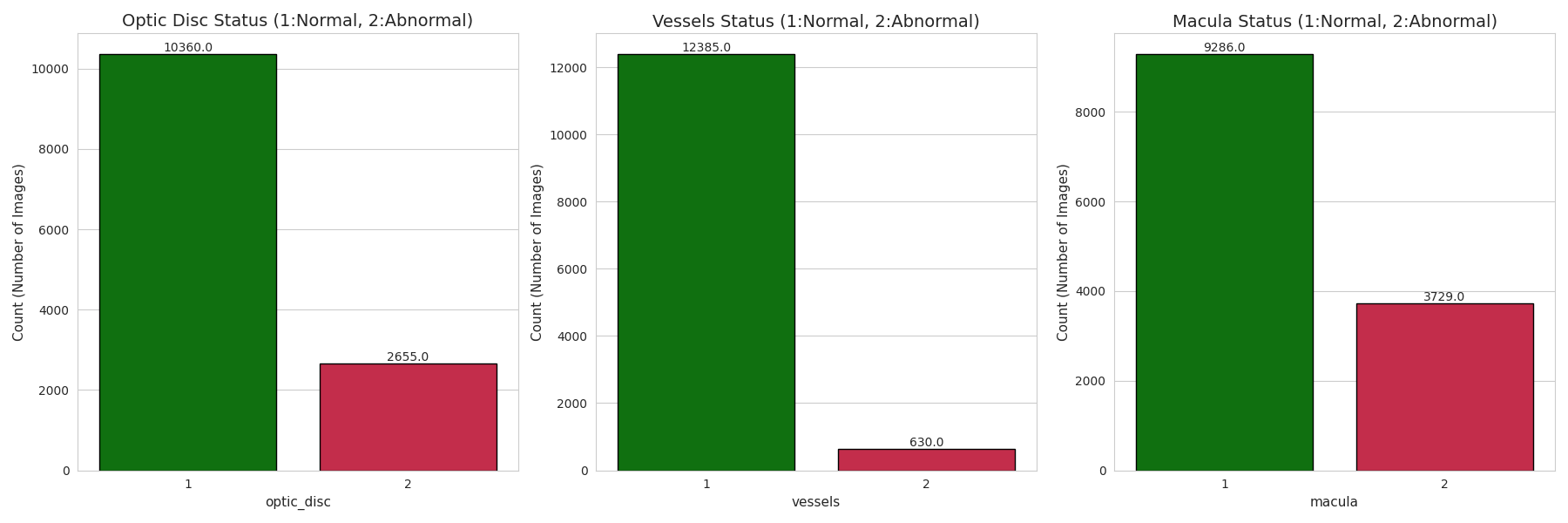}
    \caption{Anatomical status distribution}
    \label{fig:anatomical}
\end{subfigure}

\vspace{0.5em}

\begin{subfigure}{\columnwidth}
    \centering
    \includegraphics[width=\linewidth]{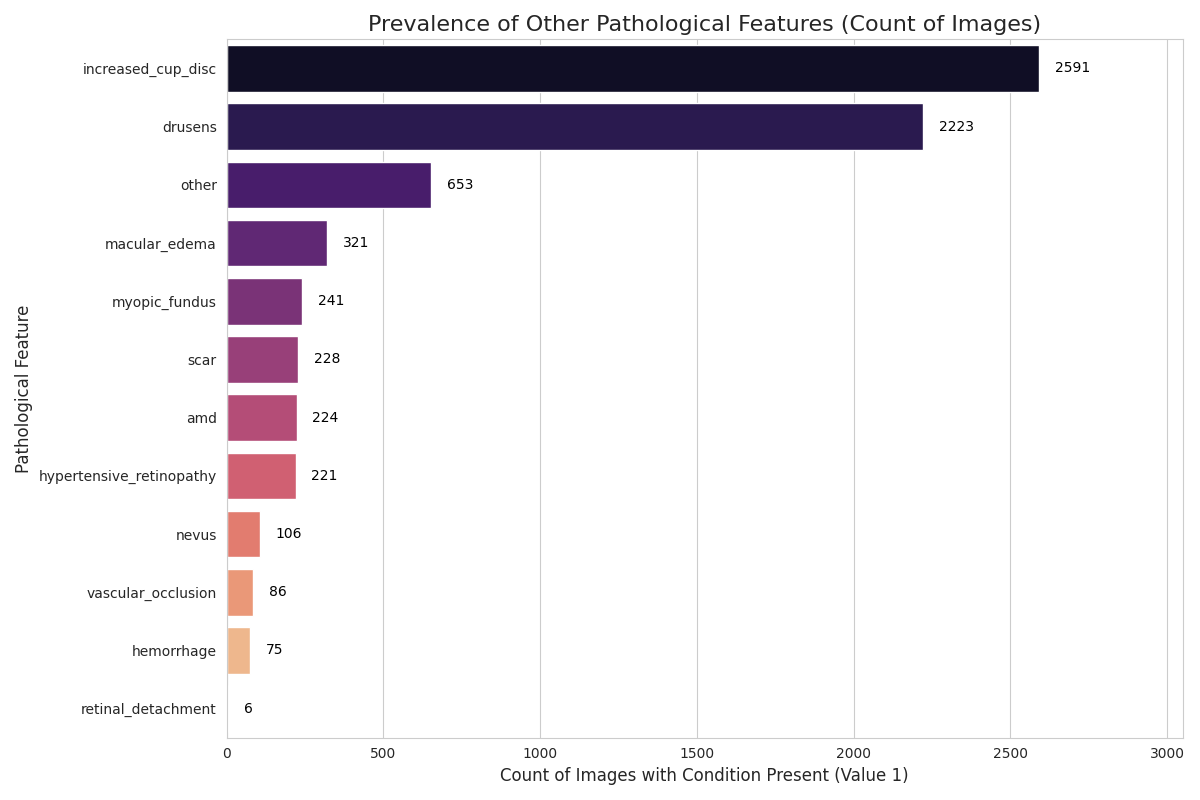}
    \caption{Other pathological features prevalence}
    \label{fig:other}
\end{subfigure}

\caption{Prevalence of pathological conditions in BRSET dataset beyond diabetic retinopathy, including macular edema, age-related macular degeneration, hypertensive retinopathy, and other retinal conditions.}
\label{fig:combined_pathology}
\end{figure}

\begin{figure}[htbp]
\centerline{\includegraphics[width=\columnwidth]{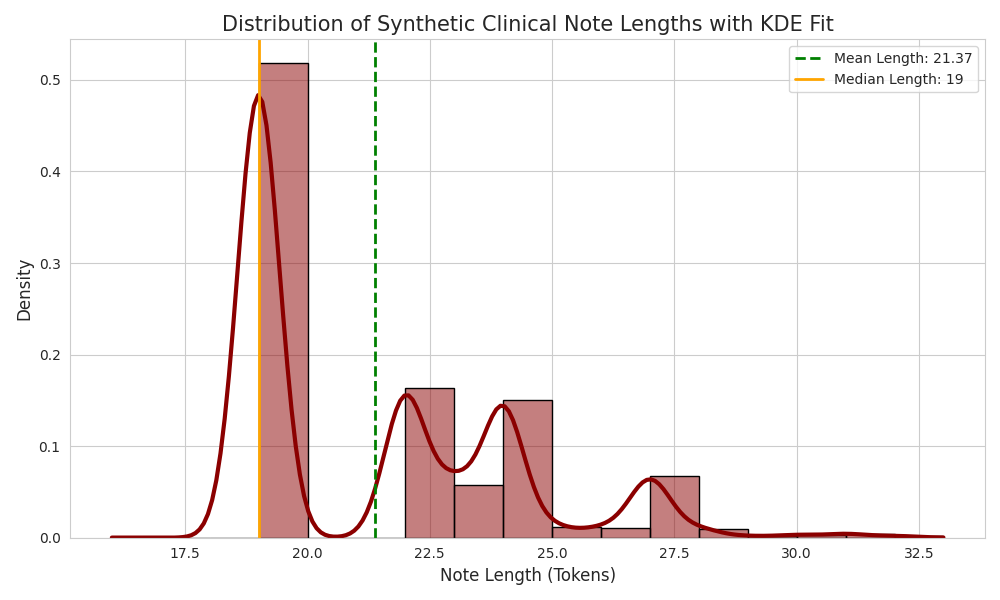}}
\caption{Distribution of synthetic clinical note lengths (in tokens) after generation from binary disease indicators. Each note falls well within the 128-token limit, validating the truncation strategy.}
\label{fig:notes}
\end{figure}

Since BRSET does not contain free-text notes, we generated synthetic clinical narratives from the dataset's binary disease indicators. For each pathological label marked as present, a short declarative sentence (e.g., \textit{``Diabetic retinopathy is present.''}) was appended to form a pseudo-clinical note, capturing diagnostic semantics consistent with ophthalmic reporting. These text strings were tokenized, truncated to 128 tokens, and embedded using Bio-ClinicalBERT.

The data were split at the patient level into 80\% training, 10\% validation, and 10\% test sets to prevent leakage. Each resulting sample comprised a preprocessed fundus image, a six-dimensional structured feature vector, and a synthetic clinical text, which together formed the multimodal input to the proposed transformer model.

\subsection{Model Architecture}

We propose a multimodal transformer framework that integrates retinal fundus images, structured clinical data, and free-text clinical notes into a unified embedding space for joint embedding creation. The model combines pretrained encoders with a modality-aware fusion transformer and multi-objective training involving contrastive alignment, modality reconstruction, and supervised classification.

\subsubsection{Vision Encoder}
The visual modality is processed using a pretrained Vision Transformer (ViT-B/16) from the timm library \cite{dosovitskiy2020image}. Each input fundus image is resized to $224 \times 224$ and tokenized into non-overlapping $16 \times 16$ patches, resulting in a sequence of patch embeddings appended with a class (CLS) token. To preserve pretrained representations while allowing domain adaptation, the initial 10 transformer blocks of the ViT are frozen, and the remaining two layers are fine-tuned. Each resulting token embedding, projected to a 256-dimensional space via a linear layer, represents the attended respective patch image descriptor. A learnable image-specific CLS token is prepended, yielding a 197-token sequence used for multimodal fusion.

\begin{figure}[htbp]
\centerline{\includegraphics[width=\columnwidth]{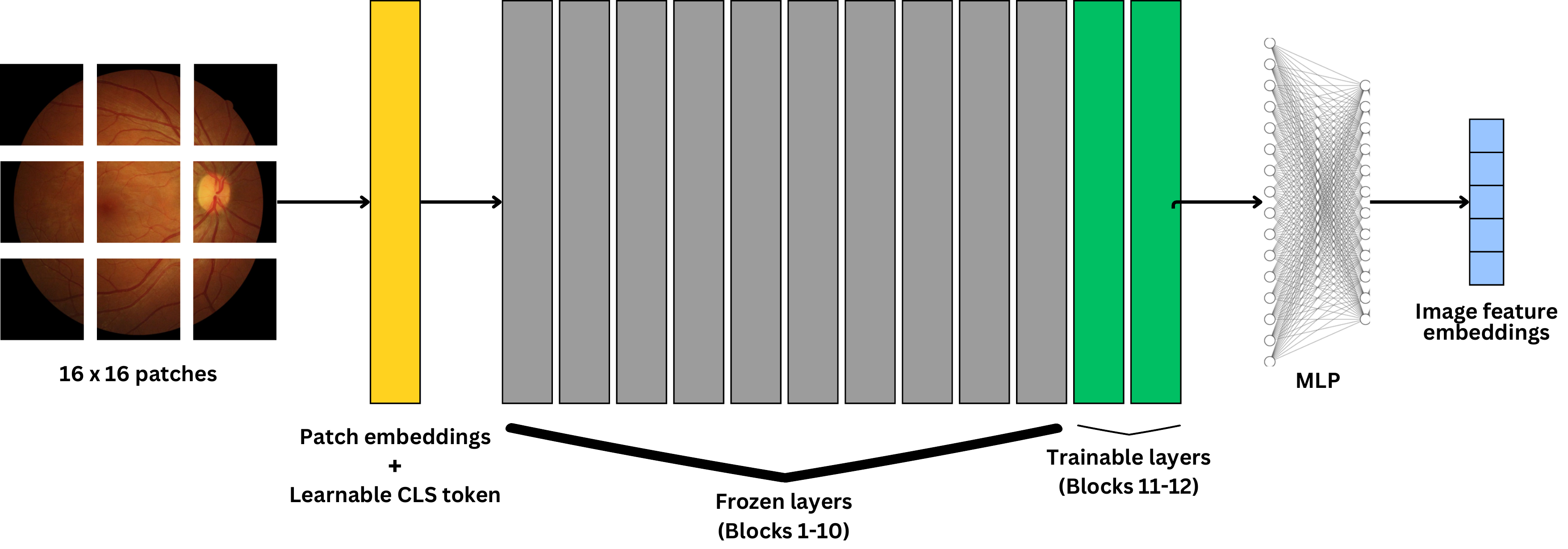}}
\caption{Overview of the ViT-B/16 architecture, followed by a fully connected layer, applied on 16 x 16 retinal image patches.}
\label{fig:vit}
\end{figure}

\subsubsection{Text Encoder}
Clinical narratives are encoded using Bio-ClinicalBERT, a domain-adapted BERT model \cite{alsentzer2019publiclyavailableclinicalbert}. Tokenized notes (maximum length 128) are passed through the model, and the contextual hidden states are projected from 768 to 256 dimensions. To reduce computational overhead, only the first 50 token embeddings are retained. As with the vision encoder, the embedding layer and the first ten transformer blocks are frozen to maintain linguistic priors. The CLS representation and truncated token embeddings together form the textual sequence.

\subsubsection{Structured Data Encoder}
Structured clinical features, which include age, sex, laterality, diabetes duration, insulin use, and diabetes type, are processed by a lightweight multilayer perceptron. The network consists of three fully connected layers with ReLU activation and dropout, mapping the six-dimensional input to a 256-dimensional latent vector. A modality-specific learnable CLS token is prepended to form the structured sequence.

\subsubsection{Multimodal Fusion Transformer}
The encoded sequences from all three modalities are concatenated and added with learnable modality-type embeddings that identify image, text, and structured tokens. This composite sequence is fed to a transformer encoder comprising six layers, eight attention heads, and a model dimension of 256. The fusion transformer enables cross-modal contextualization by allowing attention to flow across modalities at the token level. The CLS tokens from each modality, corresponding to the first tokens of their respective segments, are extracted from the output sequence and concatenated. A linear projection reduces the combined vector to a shared 256-dimensional joint embedding.

\begin{figure}[htbp]
\centerline{\includegraphics[width=0.4\columnwidth]{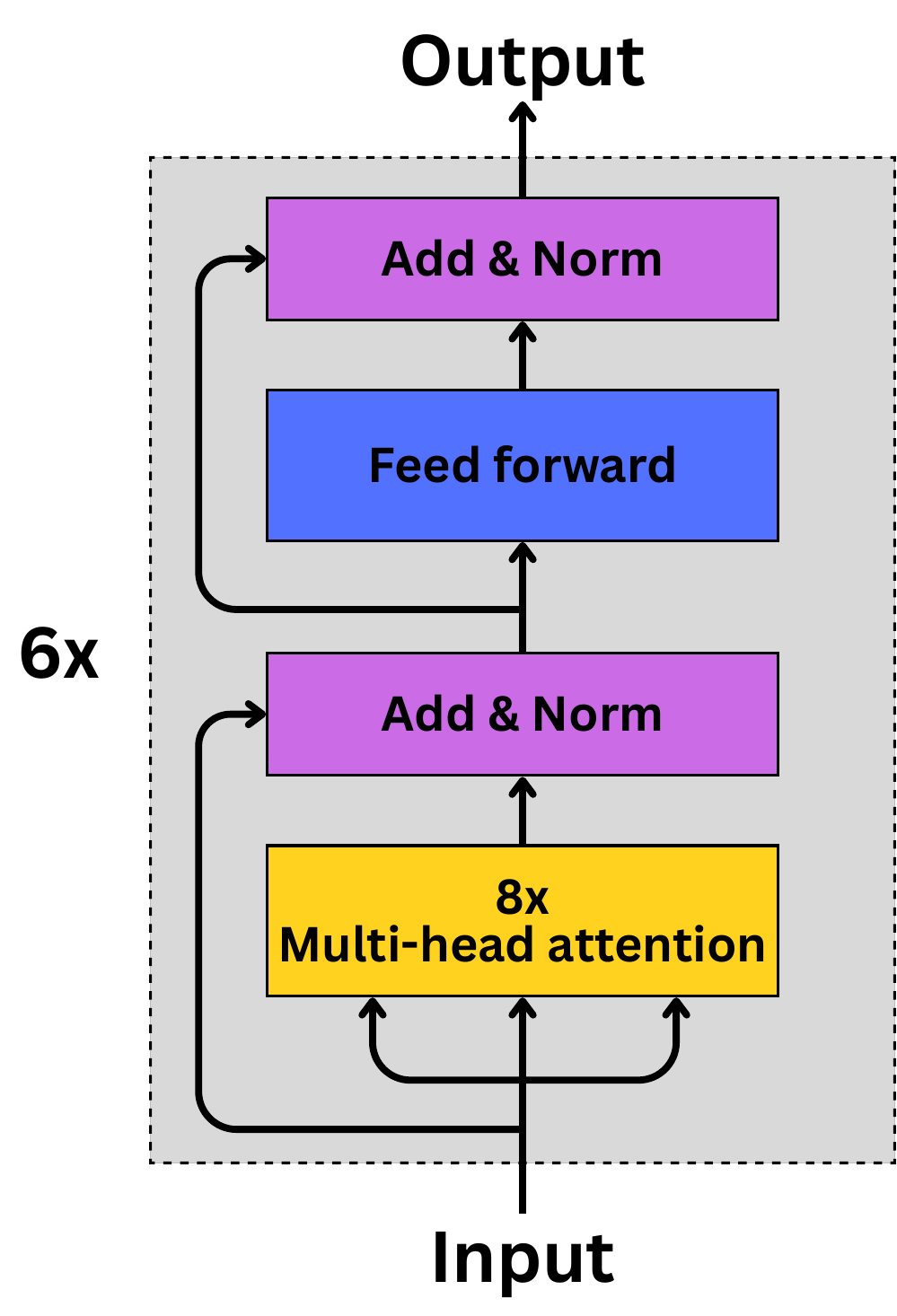}}
\caption{Overview of the multimodal fusion transformer encoder comprising six layers, eight attention heads, and a model dimension of 256, enabling cross-modal attention flow.}
\label{fig:encoder}
\end{figure}

\begin{figure*}[htbp]
\centerline{\includegraphics[width=\textwidth]{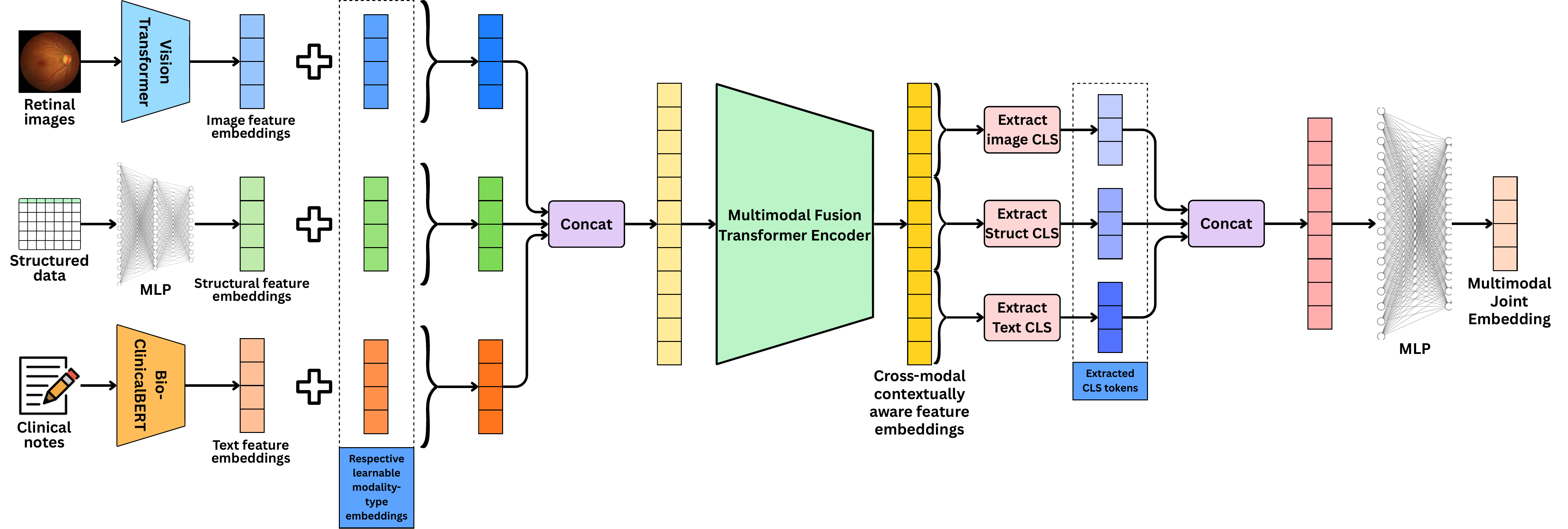}}
\caption{Overview of the proposed multimodal transformer framework showing the integration of vision, text, and structured data encoders through a fusion transformer enabling cross-modal attention flow.}
\label{fig:architecture}
\end{figure*}

\subsubsection{Decoders and Output Heads}
To regularize the modality embeddings and encourage information preservation, two reconstruction branches are incorporated:

\begin{enumerate}
\item \textbf{Image Decoder:} a convolutional decoder that progressively upsamples the image CLS embedding to reconstruct a $224 \times 224 \times 3$ image using transposed convolutions.
\item \textbf{Text Decoder:} an autoregressive transformer decoder conditioned on the text CLS embedding, trained to reconstruct the original token sequence.
\end{enumerate}

On top of the fused embedding, two linear classification heads predict the diabetic retinopathy severity under the SDRG and ICDR grading schemes, each producing five-class logits.

\subsection{Multi-Task Objective}
Training optimizes six loss components: three pairwise contrastive losses aligning image-text, image-structured, and text-structured embeddings; two reconstruction losses for image and text decoding; and a supervised classification loss averaged over SDRG and ICDR predictions. A learnable softmax-normalized weight vector dynamically balances these objectives. The total loss is thus:
\begin{equation}
\mathcal{L}_{\text{total}} = \sum_{i=1}^{6} w_i \mathcal{L}_i, \quad w_i = \frac{e^{\alpha_i}}{\sum_j e^{\alpha_j}},
\end{equation}
where $\alpha_i$ are trainable parameters. This formulation allows the model to adaptively adjust the contribution of each task during optimization.

\subsection{Implementation Details}
All embeddings share a dimension of 256. The joint transformer uses ReLU activations and layer normalization, trained with AdamW (learning rate $1\times10^{-4}$, weight decay 0.01) on NVIDIA Tesla P100-PCIE-16GB \cite{loshchilov2019decoupledweightdecayregularization}. Gradient norms are clipped to 1.0, and learning rate scheduling follows a \textit{ReduceLROnPlateau} policy. The model is trained for up to 50 epochs with early stopping based on validation loss.

\begin{figure*}[htbp]
\centerline{\includegraphics[width=\textwidth]{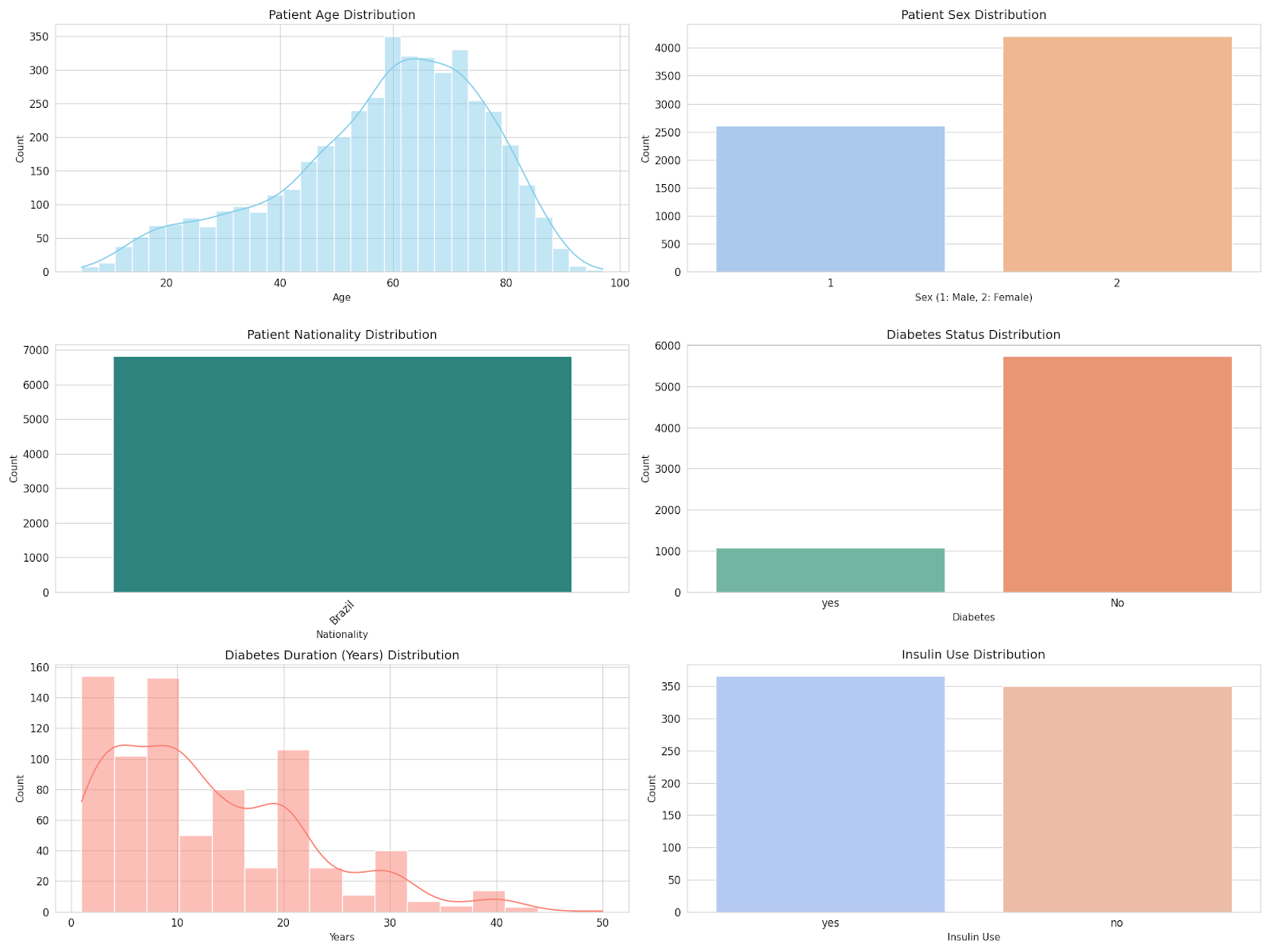}}
\caption{Overview of the patient demographic distributions in the BRSET training dataset.}
\label{fig:demographics}
\end{figure*}

\section{Experiments and Results}

\subsection{Experimental Setup}

\subsubsection{Implementation Details}
All experiments were conducted using PyTorch 1.13 on NVIDIA Tesla P100-PCIE-16GB GPUs \cite{paszke2017automatic}. The model was trained using the AdamW optimizer with a learning rate of $1 \times 10^{-4}$ and weight decay of 0.01. Gradient norms were clipped to 1.0 to ensure training stability, and learning rate scheduling followed a \textit{ReduceLROnPlateau} policy with a patience of 5 epochs and reduction factor of 0.5. Training was performed for up to 50 epochs with early stopping based on validation loss (patience of 10 epochs). The batch size was set to 8 to accommodate GPU memory constraints while maintaining effective batch statistics for contrastive learning.

The vision encoder utilized a pretrained ViT-B/16 from the timm library, with the first 10 transformer blocks frozen to preserve pretrained representations. The text encoder employed Bio-ClinicalBERT with the embedding layer and first 10 transformer blocks frozen to maintain domain-specific linguistic priors. All modality embeddings were projected to a shared 256-dimensional space. The joint transformer comprised 6 layers with 8 attention heads, ReLU activations, and layer normalization. The temperature parameter $\tau$ for contrastive losses was set to 0.07.

Data augmentation for training images included random horizontal and vertical flips, rotations of $\pm15^\circ$, and mild brightness and contrast jitter. Validation and test sets were left unaugmented to ensure consistent evaluation. The dataset was split at the patient level into training (80\%), validation (10\%), and test (10\%) sets, ensuring no patient appeared in multiple splits to prevent data leakage.

\begin{figure}[htbp]
\centerline{\includegraphics[width=\columnwidth]{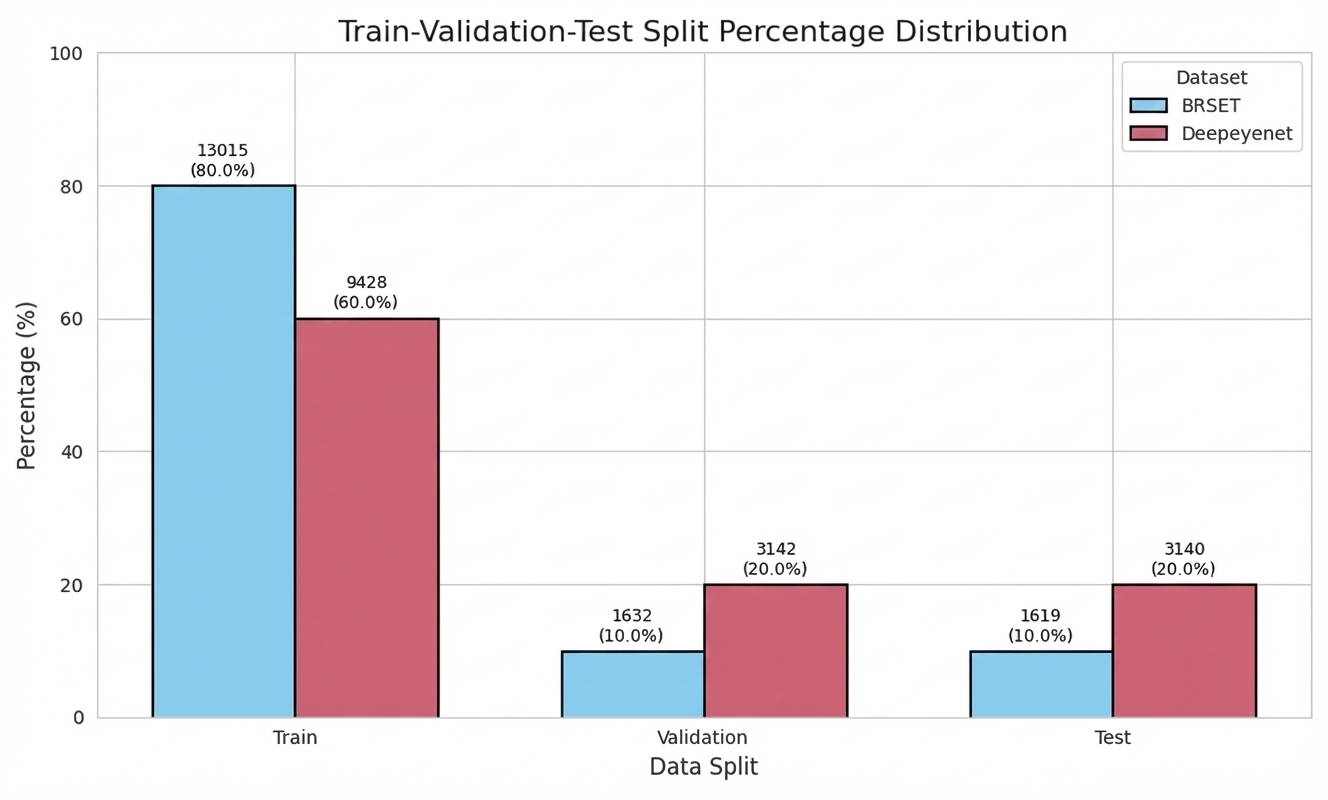}}
\caption{Patient-wise train-validation-test split to prevent any information leakage.}
\label{fig:split}
\end{figure}

\subsubsection{Evaluation Metrics}
We evaluate our model using two primary sets of metrics:

\textbf{Cross-Modal Retrieval Metrics:} For text-to-image retrieval tasks, we employ Recall@K (R@K) where K $\in$ \{1, 5, 10\}. Given a text query, R@K measures the percentage of queries for which the correct image appears in the top K retrieved results. This metric directly assesses the quality of cross-modal alignment in the learned embedding space.

\textbf{Classification Metrics:} For DR severity grading, we report classification accuracy for both SDRG and ICDR grading schemes. Accuracy is computed as the percentage of correctly classified samples across all five severity grades (0-4).

\subsubsection{Baseline Models}
We compare our approach against the following baselines:

\begin{itemize}
\item \textbf{CLIP (Zero-shot):} The standard pretrained CLIP (ViT-B/32) model without any fine-tuning, representing the off-the-shelf performance of general-domain vision-language models.

\item \textbf{CLIP (Fine-tuned):} CLIP model fine-tuned on the BRSET training set with image-text pairs, representing the best achievable performance from adapting general-domain models to the medical domain.

\item \textbf{Vision-Only Baseline:} A ViT-B/16 model trained solely on fundus images for DR classification, representing state-of-the-art unimodal approaches.

\item \textbf{SOTA Classification Models:} Published state-of-the-art results on BRSET dataset for DR severity grading \cite{restrepo2024dfdm, restrepo2024multimodaldeeplearninglowresource}.
\end{itemize}

\subsection{Cross-Modal Retrieval Performance}

Table~\ref{tab:retrieval} presents the text-to-image retrieval performance across different models. The results reveal a dramatic performance gap between general-domain vision-language models and our proposed multimodal framework.

\begin{table}[htbp]
\caption{Text-to-Image Retrieval Performance (Recall@K)}
\begin{center}
\begin{tabular}{|l|c|c|c|}
\hline
\textbf{Model} & \textbf{R@1 (\%)} & \textbf{R@5 (\%)} & \textbf{R@10 (\%)} \\
\hline
CLIP (Zero-shot) & 0.00 & 0.25 & 0.74 \\
CLIP (Fine-tuned) & 1.29 & 4.97 & 8.11 \\
\textbf{Ours (Proposed)} & \textbf{99.94} & \textbf{100.00} & \textbf{100.00} \\
\hline
\end{tabular}
\label{tab:retrieval}
\end{center}
\end{table}

\begin{figure}[htbp]
\centerline{\includegraphics[width=\columnwidth]{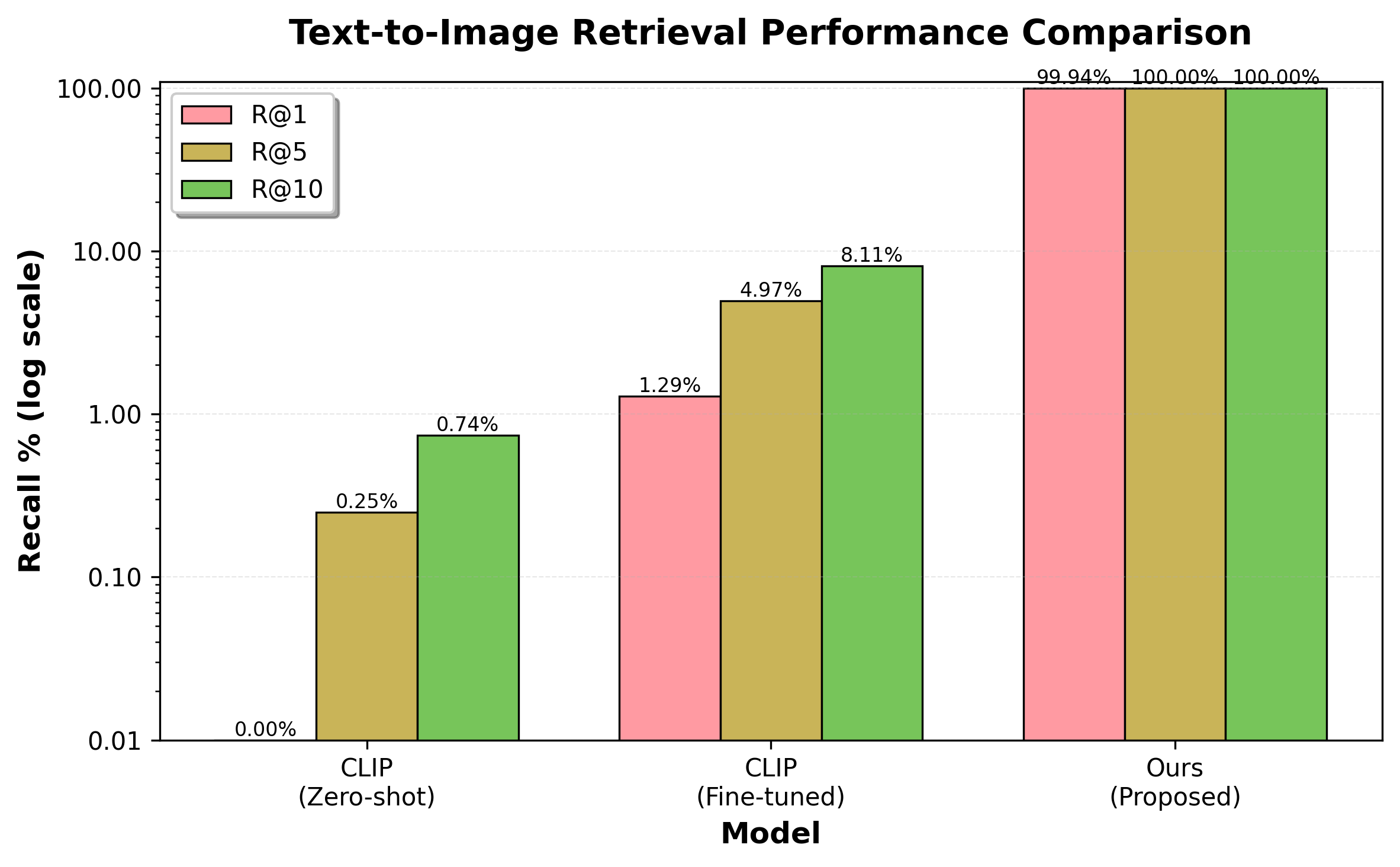}}
\caption{Text-to-Image retrieval performance (Recall@K) comparison across models. Our proposed framework achieves near-perfect retrieval while CLIP models fail catastrophically in the medical domain.}
\label{fig:retrieval}
\end{figure}

\begin{figure*}[htbp]
\centerline{\includegraphics[width=\textwidth]{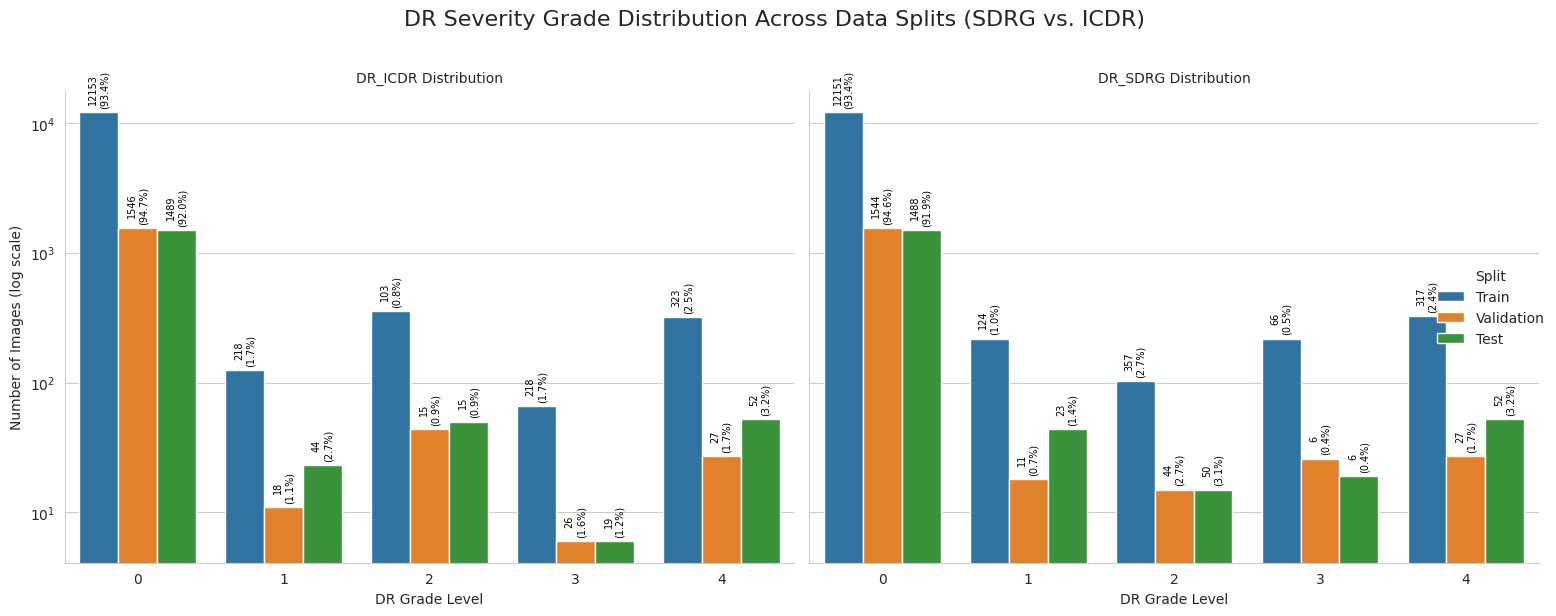}}
\caption{DR severity grade distribution across train, validation, and test splits for both SDRG and ICDR schemes, demonstrating balanced patient-level stratification to maintain similar class proportions.}
\label{fig:dr_splitwise}
\end{figure*}

The zero-shot CLIP model completely fails at the retrieval task, achieving 0\% R@1, which indicates that not a single text query successfully retrieved the correct image as the top result. This demonstrates that general-domain vision-language models lack the specialized knowledge required to align clinical descriptions with retinal pathology. Even with domain-specific fine-tuning on 11,386 BRSET training samples, CLIP's performance improves only marginally to 1.29\% R@1, 4.97\% R@5, and 8.11\% R@10.

In stark contrast, our proposed multimodal framework achieves 99.94\% R@1, with perfect 100\% performance at R@5 and R@10. This represents a 77-fold improvement over fine-tuned CLIP at R@1 and validates our hypothesis that explicitly modeling the relationships between visual features, clinical text, and structured patient data through joint transformer fusion enables effective cross-modal alignment in the medical domain.

The near-perfect retrieval performance indicates that our model has learned to map clinical descriptions such as "Diabetic retinopathy is present. Macular edema is present." to the corresponding retinal images containing microaneurysms, exudates, and edema with high precision. This capability is essential for clinical decision support systems where physicians need to retrieve relevant cases based on textual queries or generate accurate clinical reports from retinal examinations.

\subsection{Diagnostic Classification Performance}

While achieving superior retrieval performance, it is critical that the model maintains competitive diagnostic accuracy for clinical utility. Table~\ref{tab:classification} compares our model's DR severity classification performance against baseline approaches.

\begin{table}[htbp]
\caption{Diabetic Retinopathy Classification Accuracy}
\begin{center}
\begin{tabular}{|l|c|c|}
\hline
\textbf{Model} & \textbf{SDRG Acc. (\%)} & \textbf{ICDR Acc. (\%)} \\
\hline
CLIP (Zero-shot) & 97.73 & 97.97 \\
Vision-Only Baseline & 93.80 & 93.80 \\
SOTA (Binary DR) & - & 98.70 \\
\textbf{Ours (Proposed)} & \textbf{97.05} & \textbf{97.97} \\
\hline
\end{tabular}
\label{tab:classification}
\end{center}
\end{table}

\begin{figure}[htbp]
\centerline{\includegraphics[width=\columnwidth]{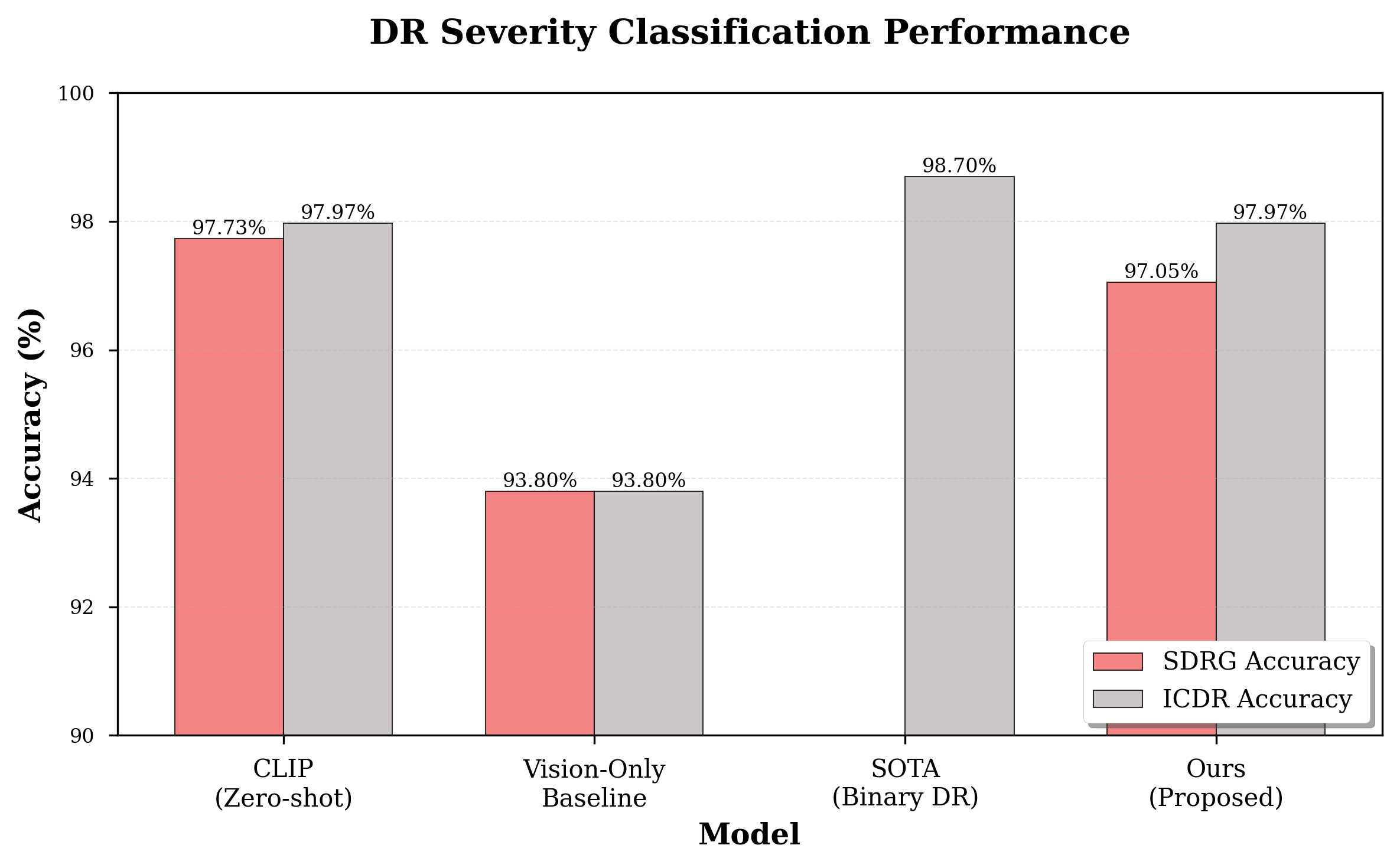}}
\caption{Diabetic Retinopathy severity classification accuracy comparison across SDRG and ICDR grading schemes. Our model maintains state-of-the-art performance while achieving superior retrieval capabilities.}
\label{fig:classification}
\end{figure}

Our model achieves 97.05\% accuracy on the SDRG grading scheme and 97.97\% on the ICDR scheme, matching the fine-tuned CLIP baseline on ICDR and remaining highly competitive with state-of-the-art classification models. The ICDR accuracy of 97.97\% ties with CLIP's performance, demonstrating that our multi-objective training strategy does not compromise diagnostic capability despite the additional complexity of learning cross-modal alignments and reconstruction objectives.

The SDRG accuracy of 97.05\% represents only a 0.68 percentage point decrease compared to fine-tuned CLIP (97.73\%), which is a negligible trade-off considering the dramatic 77-fold improvement in retrieval performance. This result validates that the reconstruction losses and contrastive alignment objectives serve as effective regularizers that enhance representation quality without sacrificing discriminative power for classification.

Notably, while CLIP achieves high classification accuracy, it does so without learning meaningful cross-modal correspondences, essentially treating the text encoder as a label embedding mechanism rather than a true semantic alignment. In contrast, our model learns rich joint representations that capture both the visual-semantic relationships necessary for retrieval and the discriminative features required for accurate diagnosis.

\subsection{Ablation Studies}

To understand the contribution of individual components, we conducted comprehensive ablation studies examining the impact of different modalities and training objectives.

\subsubsection{Modality Contribution Analysis}

\begin{table}[htbp]
\caption{Ablation Study: Impact of Different Modalities}
\begin{center}
\begin{tabular}{|l|c|c|c|}
\hline
\textbf{Configuration} & 
\makecell{\textbf{R@1 (\%)} \\ (Text-to-Image)} & 
\textbf{SDRG (\%)} & 
\textbf{ICDR (\%)} \\
\hline
Image + Text & 98.57 & 96.41 & 96.47 \\
Image + Structured & - & 94.36 & 94.24 \\
Text + Structured & - & 93.18 & 94.86 \\
\textbf{All Three (Ours)} & \textbf{99.94} & \textbf{97.05} & \textbf{97.97} \\
\hline
\end{tabular}
\label{tab:ablation_modality}
\end{center}
\end{table}

\begin{figure*}[htbp]
\centerline{\includegraphics[width=\textwidth]{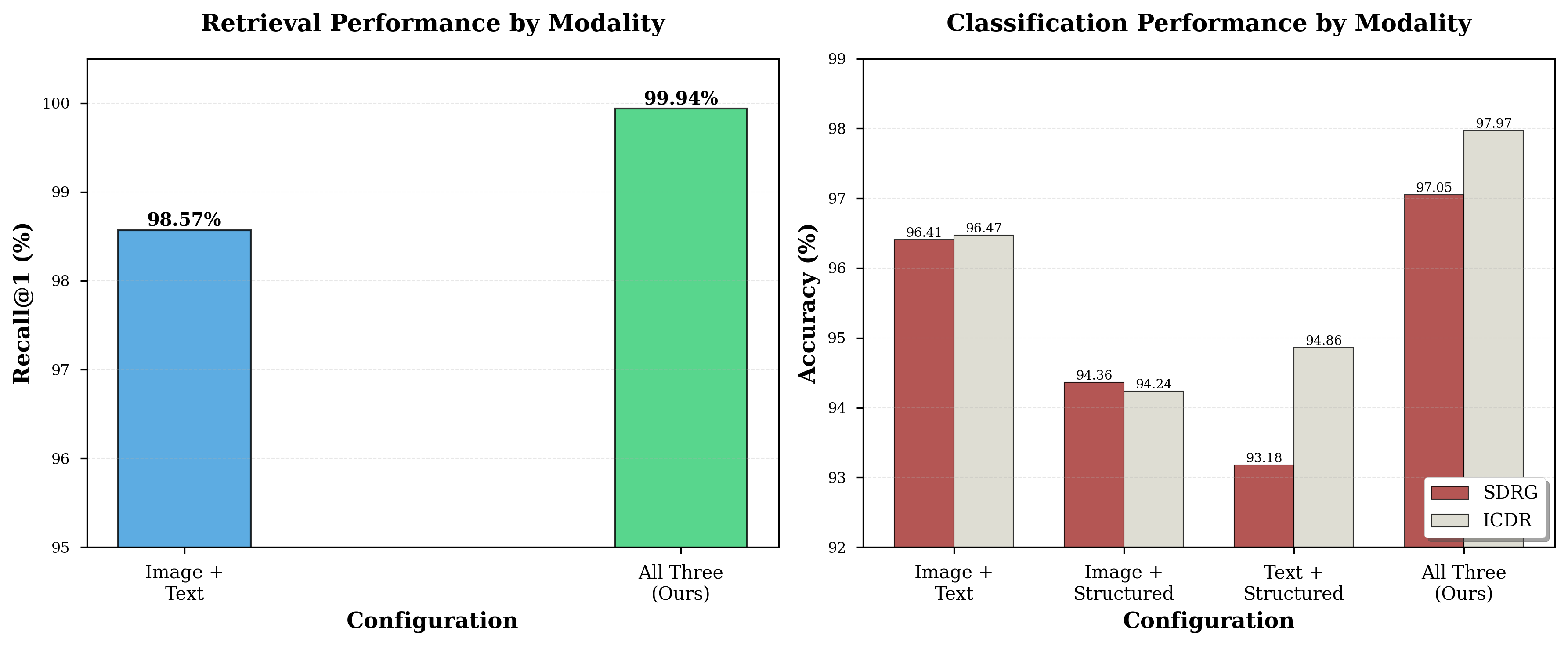}}
\caption{Impact of different modality pairs (image-text, image-structured, and text-structured) on text-to-image retrieval and Diabetic Retinopathy severity classification performance. The results confirm that the presence of both visual and textual modalities with structured data serving as a complementary role achieves the best performance.}
\label{fig:ablation_mod}
\end{figure*}

The results demonstrate that all three modalities contribute meaningfully to final performance. The image-text combination achieves strong retrieval (98.57\% R@1) but benefits from structured data integration, which improves R@1 by 1.37 percentage points to near-perfect 99.94\%. The structured data acts as a "bridge" that stabilizes alignment between the high-dimensional image space and semantic text space by providing explicit patient context (age, diabetes duration, insulin use) that correlates with disease severity.

As expected, though image-structured and text-structured pairs achieve reasonable classification accuracy (94.36\% and 93.18\% SDRG respectively), they could not outperform the image-text pair (96.41\%). On the other hand, in the absence of structured data, the image-text pair fell shorter in achieving full model's performance by 0.64\%, confirming that effective cross-modal information flow requires the presence of both visual and textual modalities with structured data serving a complementary role.

\subsubsection{Loss Function Analysis}

\begin{table}[htbp]
\caption{Ablation Study: Impact of Different Loss Components}
\begin{center}
\begin{tabular}{|l|c|c|c|}
\hline
\textbf{Loss Configuration} & \textbf{R@1 (\%)} & \textbf{SDRG (\%)} & \textbf{ICDR (\%)} \\
\hline
Classification Only & 0.19 & 95.11 & 94.92 \\
+ Contrastive & 98.33 & 96.03 & 97.03 \\
+ Reconstruction & 99.60 & 96.10 & 96.96 \\
\textbf{Full (Ours)} & \textbf{99.94} & \textbf{97.05} & \textbf{97.97} \\
\hline
\end{tabular}
\label{tab:ablation_loss}
\end{center}
\end{table}

\begin{figure}[htbp]
\centerline{\includegraphics[width=\columnwidth]{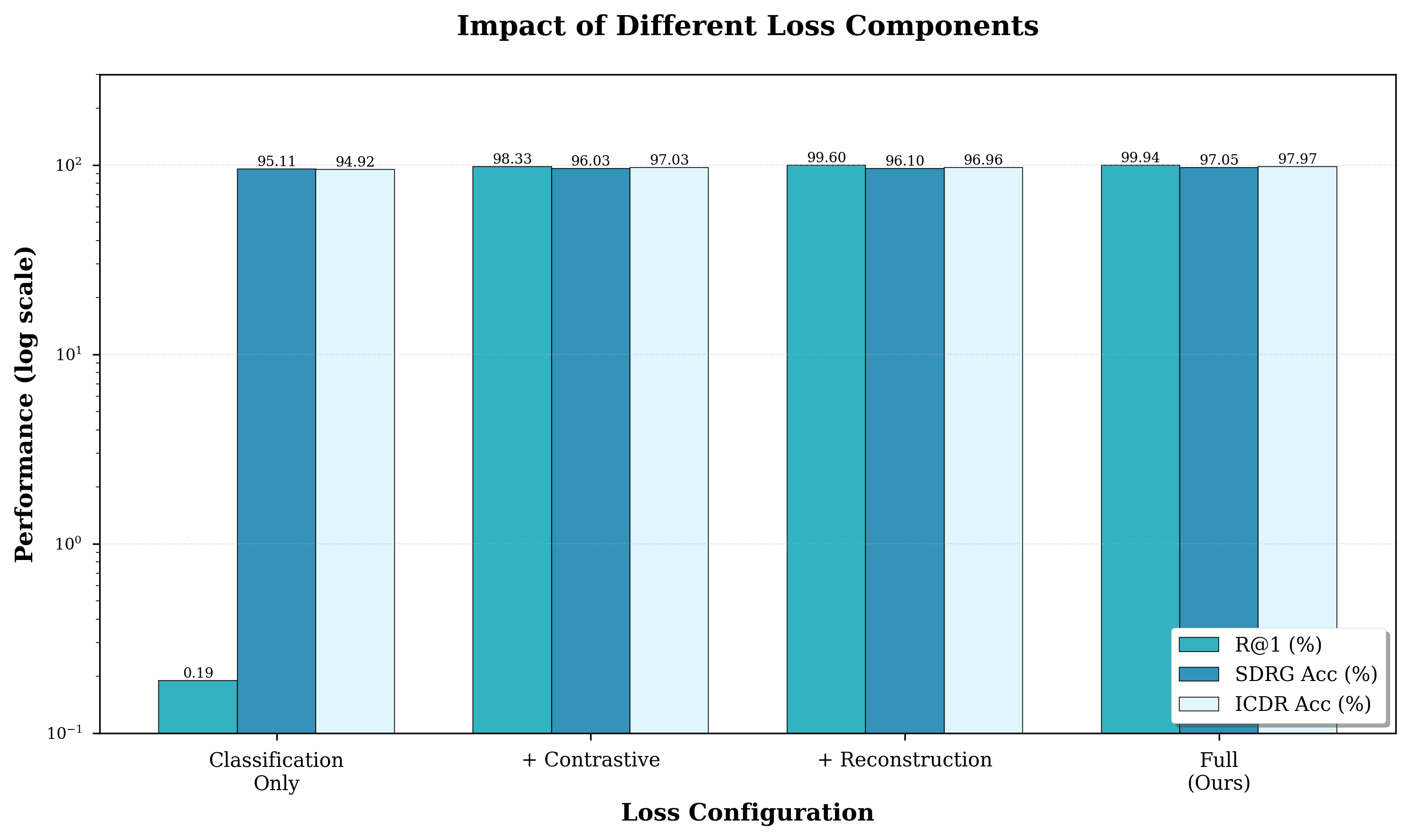}}
\caption{Impact of different loss functions (CE, CE + contrastive, and CE + contrastive + reconstruction) on text-to-image retrieval and Diabetic Retinopathy severity classification performance.  The full model with learnable loss weighting achieves the best balance, reaching 99.94\% R@1 while maintaining 97.97\% ICDR accuracy.}
\label{fig:classification}
\end{figure}

Training with classification loss alone achieves high diagnostic accuracy (95.11\% SDRG, 94.92\% ICDR) but fails at retrieval (0.19\% R@1), similar to CLIP's behavior. Adding contrastive losses dramatically improves retrieval to 98.33\% R@1, confirming that explicit cross-modal alignment objectives are essential. The reconstruction losses further boost retrieval to 99.60\% by encouraging the embeddings to retain fine-grained information necessary for regenerating original inputs. The full model with learnable loss weighting achieves the best balance, reaching 99.94\% R@1 while maintaining 97.97\% ICDR accuracy.

\subsection{Cross-Dataset Generalization}

To assess the robustness and generalizability of learned representations, we evaluated both CLIP and our model on DeepEyeNet, a completely unseen ophthalmological dataset with different image characteristics and patient demographics \cite{roy2025deepeyenetadaptivegeneticbayesian}.

\begin{table}[htbp]
\caption{Zero-Shot Performance on Unseen DeepEyeNet Dataset}
\begin{center}
\begin{tabular}{|l|c|}
\hline
\textbf{Model} & \textbf{R@1 (\%)} \\
\hline
CLIP (Fine-tuned on BRSET) & 0.22 \\
\textbf{Ours (Trained on BRSET)} & \textbf{93.95} \\
\hline
\end{tabular}
\label{tab:generalization}
\end{center}
\end{table}

\begin{figure}[htbp]
\centerline{\includegraphics[width=\columnwidth]{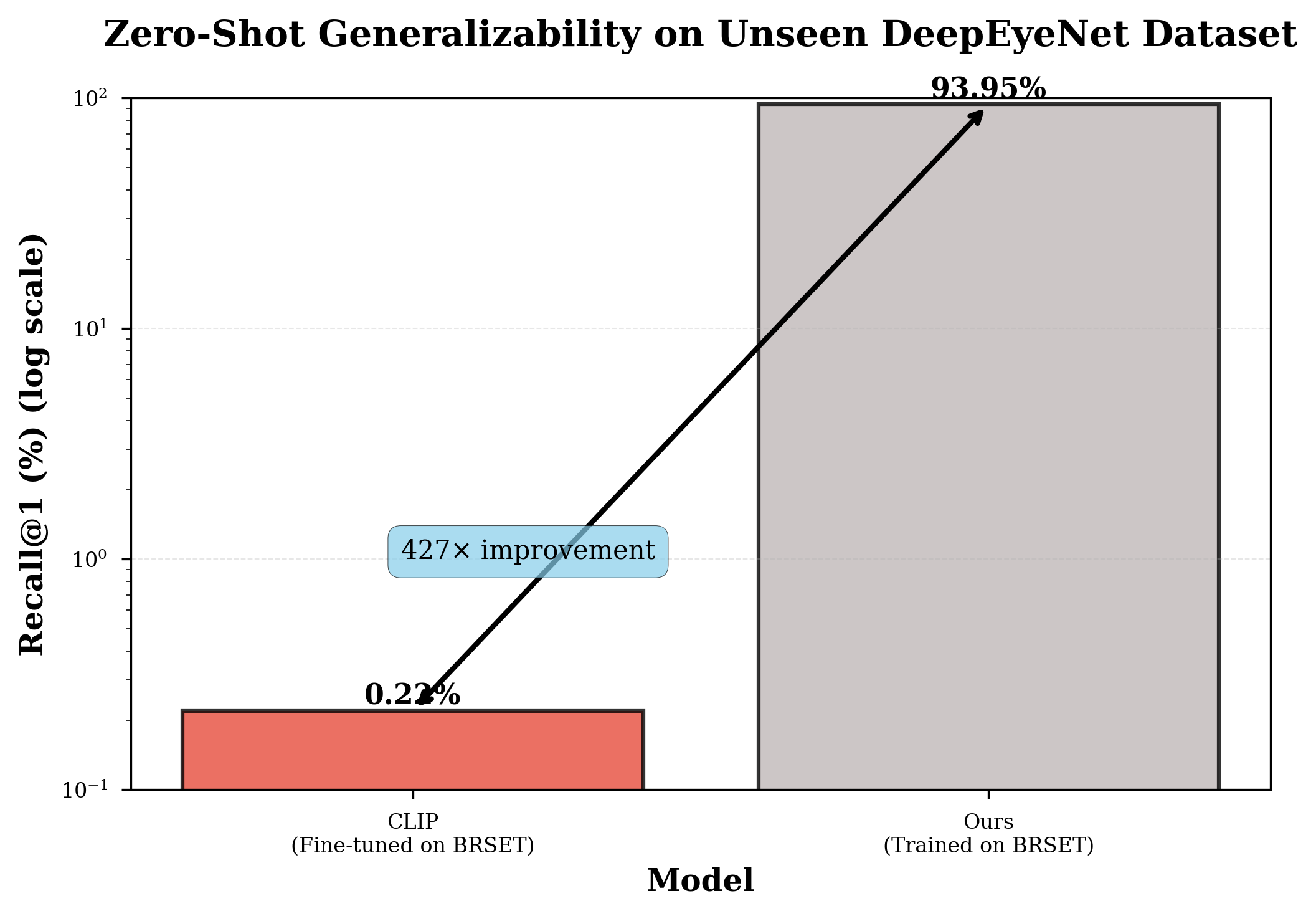}}
\caption{Zero-shot generalizability (Recall@1) on the unseen DeepEyeNet dataset. Our model demonstrates robust cross-dataset transfer while CLIP fails on unseen medical data.}
\label{fig:generalization}
\end{figure}

The results reveal a striking 427-fold improvement in cross-dataset generalization. Fine-tuned CLIP achieves only 0.22\% R@1 on DeepEyeNet, indicating that its learned representations are brittle and fail to transfer to new datasets. In contrast, our model maintains 93.95\% R@1 on completely unseen data, demonstrating that the joint embedding space has captured generalizable medical concepts rather than dataset-specific artifacts.

This strong generalization capability can be attributed to several factors: (1) the multi-objective training strategy prevents overfitting by requiring the model to simultaneously satisfy alignment, reconstruction, and classification objectives; (2) the integration of structured clinical data provides domain-invariant patient context that transfers across datasets; (3) the use of pretrained domain-adapted encoders (ViT and ClinicalBERT) with frozen early layers preserves general medical knowledge while allowing task-specific adaptation.

The cross-dataset validation establishes the practical viability of our approach for real-world clinical deployment, where models must perform reliably across diverse patient populations, imaging equipment, and clinical settings without requiring retraining for each new institution.

\subsection{Computational Performance}

\begin{table}[htbp]
\caption{Computational Requirements and Inference Time}
\begin{center}
\begin{tabular}{|l|c|c|c|}
\hline
\textbf{Model} & \textbf{Parameters} & \textbf{GPU Memory} & \textbf{Inference Time} \\
\hline
CLIP (Fine-tuned) & 151M & 4.2 GB & 23 ms \\
Vision-Only & 86M & 2.8 GB & 15 ms \\
\textbf{Ours (Proposed)} & \textbf{187M} & \textbf{6.5 GB} & \textbf{41 ms} \\
\hline
\end{tabular}
\label{tab:computational}
\end{center}
\end{table}

Our model contains 187M parameters, which is 24\% larger than fine-tuned CLIP due to the additional encoders and fusion transformer. GPU memory usage is 6.5 GB during inference, and the average inference time per sample is 41 ms on NVIDIA Tesla P100 GPUs. While computationally more demanding than unimodal baselines, the inference time remains well within acceptable ranges for clinical deployment, where screening sessions typically involve processing tens to hundreds of images rather than requiring real-time performance. The modest computational overhead is justified by the dramatic improvements in both retrieval capability and generalization performance.

\section{Conclusion}

This paper addressed a critical limitation of general-domain vision-language models in medical imaging applications: their catastrophic failure to establish meaningful cross-modal alignments between clinical descriptions and specialized medical images. Through comprehensive empirical evaluation, we demonstrated that CLIP, despite achieving high classification accuracy, fundamentally fails at text-to-image retrieval in ophthalmology, with only 1.29\% Recall@1 even after domain-specific fine-tuning. This severe misalignment limits the practical utility of such models for clinical decision support systems that require robust cross-modal reasoning capabilities.

To address this gap, we presented a knowledge-enhanced joint embedding framework that integrates retinal fundus images, clinical text narratives, and structured patient data through a multimodal transformer architecture with multi-objective training. Our approach achieved near-perfect text-to-image retrieval performance (99.94\% Recall@1), representing a 77-fold improvement over fine-tuned CLIP, while maintaining state-of-the-art diagnostic classification accuracy (97.05\% SDRG, 97.97\% ICDR). The dramatic improvement in retrieval capability without sacrificing diagnostic performance validates that explicit modeling of cross-modal relationships through joint transformer fusion and multi-task learning objectives enables effective alignment in specialized medical domains.

\subsection{Key Contributions and Findings}

Our work makes several significant contributions to multimodal medical imaging:

\textbf{Empirical Validation of Domain Gap:} We provided concrete evidence that general-domain vision-language models trained on web-scraped data fail catastrophically when applied to specialized medical contexts, achieving near-zero retrieval performance (0\% for zero-shot CLIP, 1.29\% after fine-tuning) despite maintaining high classification accuracy. This finding has important implications for the medical AI community, suggesting that off-the-shelf vision-language models require fundamental architectural and training modifications rather than simple fine-tuning for medical applications.

\textbf{Effective Multimodal Integration:} Through comprehensive ablation studies, we demonstrated that all three modalities, \textit{viz}, images, text, and structured data, contribute meaningfully to final performance. The integration of structured clinical features (age, diabetes duration, insulin use) improved retrieval from 98.57\% to 99.94\% R@1, acting as a "bridge" that stabilizes alignment between high-dimensional visual features and semantic text representations. This validates the clinical intuition that diagnostic reasoning benefits from integrating patient context alongside imaging findings.

\textbf{Multi-Objective Training Strategy:} Our analysis revealed that combining contrastive alignment losses, reconstruction losses, and classification losses with learnable dynamic weighting enables the model to learn representations that simultaneously excel at cross-modal retrieval and diagnostic classification. Training with classification loss alone yielded only 0.19\% R@1, while adding contrastive objectives improved this to 98.33\%, and incorporating reconstruction losses further enhanced performance to 99.60\%, with the full system achieving 99.94\%.

\textbf{Strong Generalization:} Perhaps most significantly, our model demonstrated robust cross-dataset generalization, maintaining 93.95\% R@1 on the completely unseen DeepEyeNet dataset compared to fine-tuned CLIP's 0.22\%. This 427-fold improvement in zero-shot transfer establishes that our approach learns generalizable medical concepts rather than dataset-specific artifacts, addressing a critical concern for real-world clinical deployment where models must perform reliably across diverse patient populations and institutional settings.

\subsection{Clinical Implications}

The capabilities demonstrated by our framework have several important implications for clinical practice:

\textbf{Clinical Decision Support:} The near-perfect retrieval performance enables physicians to efficiently query image databases using natural language descriptions of pathological findings, facilitating evidence-based decision making through rapid access to similar historical cases.

\textbf{Automated Report Generation:} The learned cross-modal alignment provides a foundation for generating accurate clinical reports from retinal examinations, potentially reducing the documentation burden on ophthalmologists while maintaining diagnostic accuracy.

\textbf{Educational Applications:} Medical students and residents can leverage the retrieval capabilities to find relevant teaching cases by describing clinical features, supporting case-based learning and diagnostic skill development.

\textbf{Resource-Limited Settings:} The strong generalization across datasets suggests potential for deploying the model in resource-limited settings where local training data may be scarce, addressing the critical shortage of specialized ophthalmologists in low- and middle-income countries.

\subsection{Limitations and Future Directions}

While our work demonstrates significant advances, several limitations warrant acknowledgment and suggest directions for future research:

\textbf{Synthetic Clinical Notes:} The BRSET dataset lacks authentic clinical notes, requiring us to generate synthetic narratives from binary disease indicators. While this approach proved effective for demonstrating the value of multimodal integration, real-world clinical notes contain richer information including temporal progression, treatment history, and physician observations that could further enhance model performance. Future work should validate our approach on datasets with authentic clinical documentation to assess generalization to natural clinical language with its inherent noise, ambiguity, and complexity.

\textbf{Computational Requirements:} Our model contains 187M parameters and requires 41 ms inference time per sample, representing a 24\% increase in parameters and 78\% increase in inference time compared to fine-tuned CLIP. While these requirements remain within acceptable ranges for clinical screening workflows, deployment on resource-constrained edge devices or real-time applications may require model compression techniques such as knowledge distillation, pruning, or quantization. Future work should explore efficient variants that maintain performance while reducing computational demands.

\textbf{Single Disease Focus:} This study focused exclusively on diabetic retinopathy severity grading. The framework's applicability to other ophthalmological conditions (age-related macular degeneration, glaucoma, retinal vein occlusion) and other medical imaging domains (radiology, pathology, dermatology) remains to be validated. Extending the approach to multi-disease scenarios and demonstrating transfer across medical specialties would establish broader clinical utility.

\textbf{Temporal Modeling:} Current implementation processes individual examinations independently, neglecting the longitudinal nature of disease progression. Incorporating temporal information through sequential modeling of patient records could enable disease progression prediction and risk stratification, providing additional clinical value beyond static classification and retrieval.

\textbf{Interpretability and Uncertainty:} While our ablation studies and attention mechanisms provide some interpretability, clinical deployment requires robust uncertainty quantification to identify cases requiring human expert review. Future work should incorporate Bayesian approaches or ensemble methods to provide calibrated confidence estimates alongside predictions.

\textbf{Prospective Clinical Validation:} All experiments were conducted retrospectively on existing datasets. Prospective validation in real clinical workflows is essential to assess practical impact on diagnostic accuracy, physician efficiency, and patient outcomes. Such studies should also evaluate potential failure modes, edge cases, and the model's behavior under distribution shift.

\textbf{Federated Learning:} To address privacy concerns and enable collaborative model improvement across institutions without data sharing, future work should explore federated learning approaches that allow training on decentralized medical datasets while preserving patient confidentiality.

\textbf{Multimodal Report Generation:} While our work establishes effective cross-modal alignment, extending the framework to generate complete clinical reports rather than simple retrieval could provide additional clinical utility. This would require incorporating natural language generation capabilities with appropriate medical domain constraints.

\subsection{Broader Impact}

This work demonstrates that specialized medical imaging applications require purpose-built architectures and training strategies rather than direct application of general-domain models. The dramatic performance gap between CLIP and our domain-specific approach (1.29\% vs. 99.94\% R@1) highlights the importance of incorporating medical domain knowledge, structured clinical data, and multi-objective training for healthcare AI systems.

The strong cross-dataset generalization demonstrated by our framework suggests potential for democratizing access to high-quality diagnostic tools in underserved regions where ophthalmologist shortages are most acute. However, deployment must be accompanied by careful consideration of ethical implications including algorithmic bias, equitable access, and appropriate human oversight.

\subsection{Final Remarks}

We have presented a comprehensive multimodal framework for diabetic retinopathy assessment that achieves near-perfect cross-modal retrieval while maintaining state-of-the-art diagnostic accuracy. By explicitly modeling relationships between retinal images, clinical text, and structured patient data through joint transformer fusion and multi-objective training, our approach addresses critical limitations of general-domain vision-language models in specialized medical contexts.

The dramatic improvements in both retrieval performance (77-fold over fine-tuned CLIP) and cross-dataset generalization (427-fold improvement) establish the viability of this approach for real-world clinical deployment. As medical AI systems continue to evolve toward multimodal reasoning capabilities that mirror human clinical decision-making, our work provides a foundation for developing robust, generalizable, and clinically useful diagnostic support systems.

All code, pretrained models, and experimental protocols will be made publicly available to facilitate reproducibility and encourage further research in multimodal medical imaging. We hope this work inspires continued investigation into purpose-built architectures for medical AI that leverage the full richness of multimodal clinical data to improve patient care worldwide.

\bibliographystyle{IEEEtran}
\bibliography{references}

\end{document}